\documentclass[cameraready]{Interspeech}
\usepackage{cite}
\usepackage{multirow}    % 表格多行合并
\usepackage{booktabs}    % 表格线（toprule等）
\usepackage{graphicx}    % 调整表格大小
\usepackage{xcolor}  % 颜色支持（带table选项）
\usepackage{colortbl}     % 表格颜色功能
\usepackage{url}
\interspeechcameraready

\title{MVCL-DAF++: Enhancing Multimodal Intent Recognition via Prototype-Aware Contrastive Alignment and Coarse-to-Fine Dynamic Attention Fusion}

\title{MVCL-DAF++: Enhancing Multimodal Intent Recognition via Prototype-Aware Contrastive Alignment and Coarse-to-Fine Dynamic Attention Fusion}

\author[affiliation={1}]{Haofeng}{Huang}
\author[affiliation={2}]{Bin}{Li$^\dagger$}
\author[affiliation={2}]{Yifei}{Han}
\author[affiliation={2}]{Long}{Zhang}
\author[affiliation={3}]{Yangfan}{He}
\author[affiliation={4}]{Yaxin}{Xue}

\affiliation{}{University of Shanghai for Science and Technology}{China}
\affiliation{}{Shenzhen Institute of Advanced Technology, Chinese Academy of Sciences}{China}
\affiliation{}{University of Minnesota-Twin Cities, USA \quad $^{4}$University of Leeds}{UK}

\email{2235060414@st.usst.edu.cn, b.li2@siat.ac.cn}


\usepackage{comment}

\begin{document}

\maketitle

\begingroup
\renewcommand{\thefootnote}{\fnsymbol{footnote}}
\footnotetext[2]{Corresponding author.}
\endgroup
% \vspace{-1.5em}
% \noindent\textsuperscript{\dagger}Corresponding author.
\begin{abstract}
    
    Multimodal intent recognition (MMIR) suffers from weak semantic grounding and poor robustness under noisy or rare-class conditions. We propose MVCL-DAF++, which extends MVCL-DAF with two key modules: (1) Prototype-aware contrastive alignment, aligning instances to class-level prototypes to enhance semantic consistency; and (2) Coarse-to-fine attention fusion, integrating global modality summaries with token-level features for hierarchical cross-modal interaction. On MIntRec and MIntRec2.0, MVCL-DAF++ achieves new state-of-the-art results, improving rare-class recognition by +1.05\% and +4.18\% WF1, respectively. These results demonstrate the effectiveness of prototype-guided learning and coarse-to-fine fusion for robust multimodal understanding.
\end{abstract}

\section{Introduction}

Intent recognition has traditionally been studied in unimodal settings, 
where user intentions are inferred from a single information source such as text or visual input. 
While these approaches have achieved strong performance in constrained environments, 
they remain fundamentally limited in real-world human–computer interaction, 
where intentions are often conveyed through multiple complementary communication channels. 
In natural conversations, users express intentions not only through linguistic content, 
but also through vocal cues such as prosody and intonation, as well as visual signals 
including facial expressions and gestures. 
Ignoring these heterogeneous signals may lead to incomplete or ambiguous understanding.

To address this limitation, multimodal intent recognition (MMIR)~\cite{b1, zhu2025survey} has emerged 
as a critical research direction, aiming to jointly model textual, acoustic, and visual 
information for more reliable intent inference. 
By integrating spoken language, facial expressions, and vocal characteristics, 
multimodal approaches enable a richer and more fine-grained understanding of user behavior, 
which is essential for building intelligent conversational agents and human-centered AI systems 
\cite{diao2025temporal, li2024towards, yu2025visual}. 
Recent benchmark datasets such as MIntRec~\cite{b5} and MIntRec2.0~\cite{b6} further highlight 
the challenges of real-world MMIR, including open-intent settings, long-tailed distributions, 
cross-modal ambiguity, and noisy inputs.
Recent research has explored diverse strategies to address these challenges. 
\textbf{Attention-based fusion} methods~\cite{b7,b18,b21} adaptively weight informative signals across modalities, 
and more recent works propose dynamic or hierarchical variants~\cite{b8,b10} to capture multi-scale interactions better. 
In parallel, \textbf{contrastive learning approaches}~\cite{b11,b13,b14,b19,b20} encourage alignment among modality-specific representations in a shared embedding space, 
with token-level contrastive objectives~\cite{b11} and contextual augmentation~\cite{b17} further enhancing robustness under ambiguous or low-resource settings. 
Beyond class-level learning, \textbf{prototype-based or semantic grounding strategies}~\cite{b12,b20} have shown effectiveness in few-shot and long-tailed recognition, 
though they remain under-explored in multimodal intent recognition. To support systematic evaluation, \textbf{benchmark datasets} such as MIntRec~\cite{b5} and MIntRec2.0~\cite{b6} 
provide realistic conversational scenarios with open-intent, imbalanced, and cross-modal ambiguous samples, 
driving progress in the community. 
A representative method, \textbf{MVCL-DAF}~\cite{b18}, combines multiview contrastive learning with dynamic attention fusion (DAF) to selectively integrate and regularize multimodal inputs. However, it still faces two critical limitations: (1) contrastive alignment is performed purely at the instance level, without \textit{explicit semantic grounding}, making the model vulnerable to noise and ambiguous samples\cite{b24, b25}; and (2) the fusion strategy treats modality inputs as flat token sequences, ignoring hierarchical semantic structures and redundancies, particularly in visual and acoustic modalities\cite{b22, b23}.
We propose \textbf{MVCL-DAF++}, an enhanced MMIR framework with two key innovations to address these limitations. First, we introduce \textit{prototype-aware contrastive alignment}, in which class-level semantic prototypes are learned and instance-to-prototype contrastive learning is performed. This enables explicit semantic grounding that improves robustness to noisy and ambiguous inputs. Second, we design a \textit{coarse-to-fine dynamic attention fusion} mechanism, where a Modality-Aware Transformer Encoder extracts coarse-grained global summaries for each modality, which are then dynamically integrated with fine-grained token-level features via the DAF module, enabling hierarchical and adaptive cross-modal interaction. \textbf{Our main contributions are as follows:} (1) \textbf{Prototype-aware contrastive alignment}: We introduce class-level prototypes to ground contrastive learning in shared semantics explicitly, improves cross-modal alignment robustness. (2) \textbf{Coarse-to-fine dynamic attention fusion}: We design a modality-aware encoder that unifies global and token-level features through dynamic attention fusion. (3) \textbf{Extensive empirical validation}: We establish new state-of-the-art performance on \textbf{MIntRec\cite{b5} and MIntRec2.0\cite{b6}}, with consistent gains in standard, rare-class, and noisy-input scenarios, demonstrating the generalizability of our approach to real-world MMIR tasks.
\begin{figure*}[htbp]
\centering
\includegraphics[width=1.0\textwidth,trim=135 95 310 110,clip]{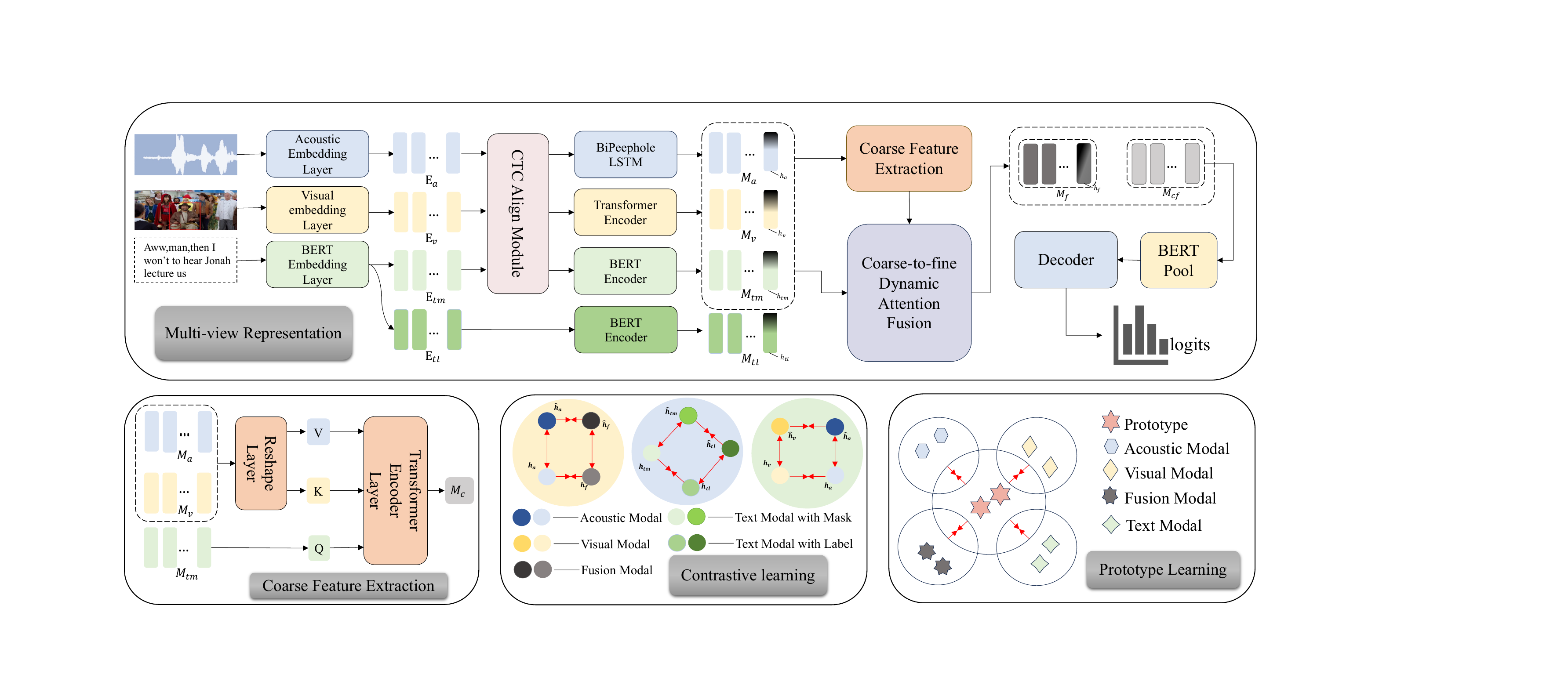}
\vspace{-1em}
\caption{MVCL-DAF++ architecture with four modules: (1) Modality encoders with coarse-to-fine DAF, 
(2) Cross-modal coarse feature extraction, (3) Contrastive learning for representation regularization, and 
(4) Prototype-aware contrastive alignment.}
\vspace{-1em}
\label{fig:architecture}
%\vspace{-1em}
\end{figure*}
\section{Method}
\vspace{-0.5em}
\subsection{Model Overview}
\vspace{-0.5em}
As illustrated in Fig.~1, the proposed \textbf{MVCL-DAF++} consists of four components: 
(1) multi-view representation learning to encode textual, visual, and acoustic inputs and coarse-to-fine DAF fusion to integrate global and token-level features; 
(2) Coarse feature extraction to capture global modality-level signals; 
(3) Representation regularization with contrastive learning; and (4) Prototype-aware contrastive alignment for robust semantic grounding. 
The final outputs are optimized with both classification and contrastive objectives.  
Following MVCL-DAF~\cite{b18}, we also retain the CTC Align Module, the BiPeephole LSTM for audio, and the decoder with BERT pooling, which are reused without modification.

Formally, let $\mathcal{X}=\{X^{t},X^{v},X^{a}\}$ denote the multimodal input, 
where $X^{m}\in\mathbb{R}^{L_m\times d_m}$ represents the sequence of $L_m$ tokens from modality $m\in\{t,v,a\}$ 
with embedding dimension $d_m$. 
After modality-specific encoders, we obtain low-level embeddings 
$E_a$ (acoustic), $E_v$ (visual), $E_{tm}$ (text-masked), and $E_{tl}$ (text-labeled). 
These are transformed into contextualized features 
$M_a, M_v, M_{tm}, M_{tl}$, which are further fused into a coarse representation $M_c$. 
For each instance $i$, its final hidden embedding is denoted as $\mathbf{h}_i$ (e.g., $\mathbf{h}_a,\mathbf{h}_v,\mathbf{h}_{tm},\mathbf{h}_{tl}, \mathbf{h}_{f}, \mathbf{h}_{cf}$). 
The fused representation $M_{cf}$ is then passed to the decoder and classifier to produce logits for intent and The target y denotes the ground-truth intent label of the input sample.

\begin{table*}[ht]
\centering
\setlength{\tabcolsep}{6pt}
\renewcommand{\arraystretch}{0.8}
\caption{Performance comparison on MIntRec\cite{b5} and MIntRec2.0\cite{b6} datasets. All results are averaged over 10 random seeds (0–9). Best results are highlighted in bold.}
\label{tab:results}
\scriptsize 
\resizebox{\linewidth}{!}{%
\begin{tabular}{l|cccc|cccc}
\toprule
\multirow{2}{*}{\textbf{Methods}}  & \multicolumn{4}{c|}{\textbf{MIntRec}\cite{b5}} & \multicolumn{4}{c}{\textbf{MIntRec2.0}\cite{b6}} \\
& ACC (\%) & WF1 (\%) & WP (\%) & R (\%) & ACC (\%) & WF1 (\%) & WP (\%) & R (\%) \\
\midrule
MulT~\cite{b7} & 72.52 & 71.80 & 72.60 & 67.44 & 56.95 & 54.26 & 54.49 & 40.65 \\
MAG-BERT~\cite{b19} & 72.16 & 71.30 & 72.03 & 67.61 & 55.87 & 52.58 & 53.71 & 39.93 \\
TCL-MAP~\cite{b11} & 73.69 & 73.38 & 73.90 & 71.59 & 56.99 & 54.33 & 55.07 & 41.87 \\
MVCL-DAF~\cite{b18} & 74.72 & 74.61 & 75.07 & 71.94 & 57.80 & 55.05 & 55.82 & 42.03 \\
\rowcolor{gray!10} 
MVCL-DAF++ (Ours) & \textbf{76.18} & \textbf{75.66} & \textbf{76.17} & \textbf{74.39} & \textbf{60.40} & \textbf{59.23} & \textbf{60.51} & \textbf{53.96} \\
\textbf{Performance Improvement} & +1.46 & +1.05 & +1.10 & +2.45 & +2.60& +4.18 &+4.69& +11.93\\
\bottomrule
\end{tabular}
}
\end{table*}
% \vspace{-1.5em}
\subsection{Prototype-Aware Contrastive Alignment}
\vspace{-0.5em}
To enhance semantic consistency, we introduce class-level prototypes as anchors. 
For each class $c$, its prototype $\mathbf{r}_{c}$ is computed within a mini-batch as
\begin{equation}
\mathbf{r}_{c}=\frac{1}{|\mathcal{I}_{c}|}\sum_{i\in\mathcal{I}_{c}}\mathbf{h}_{i},
\end{equation}
where $\mathcal{I}_{c}=\{i|y_{i}=c\}$ and $\mathbf{h}_{i}$ is the instance embedding.  
Prototypes are L2-normalized, and a prototype-aware InfoNCE loss is defined as
\begin{equation}
\mathcal{L}_{\text{proto}}
=-\log \frac{\exp(\text{sim}(\mathbf{h}_{i},\mathbf{r}_{y_i})/\tau)}
{\sum_{c=1}^{C}\exp(\text{sim}(\mathbf{h}_{i},\mathbf{r}_{c})/\tau)},
\end{equation}
where $\tau$ is a temperature hyperparameter.  
This enforces class-level structure and improves robustness against noisy and imbalanced data.  
% \vspace{-1.5em}
\subsection{Coarse-to-fine Dynamic Attention Fusion}
\vspace{-0.5em}
We next capture hierarchical interactions via a coarse-to-fine strategy.  
Given token embeddings, we compute a coarse representation $M_{c}$ 
using a modality-aware Transformer:

\vspace{-0.25em}
\begin{equation}
M_{c}=\text{Enc}(Q=Text,\;K=Visual,\;V=Acoustic),
\end{equation}

where the three modalities are assigned distinct functional roles. 
The textual features serve as semantic queries that guide the search 
for complementary cross-modal evidence, while the visual features 
provide the similarity space for relevance matching and the acoustic 
features supply additional contextual cues to be aggregated. 
This design enables the model to retrieve modality-specific information 
that complements the textual semantics and forms a coarse global 
multimodal representation $M_c$.

$M_{c}$ is then combined with token-level features through two DAF modules, 
yielding fine-grained $M_{f}$ and coarse-enhanced $M_{cf}$.  
$M_{f}$ supports contrastive learning, while $M_{cf}$ is used for classification.

% \usepackage{booktabs}   % \toprule \midrule \bottomrule
% \usepackage{amsmath,amssymb} % 提供 \checkmark 和 \times

% \vspace{-1.5em}
\subsection{Multi-View Representation Learning}
\vspace{-0.5em}
To align modalities, we adopt InfoNCE loss.  
For anchor $\mathbf{h}_{anchor}$ and positive $\mathbf{h}_{p}$:
\begin{equation}
\mathcal{L}_{\text{InfoNCE}}
=-\log \frac{\exp(\text{sim}(\mathbf{h}_{anchor},\mathbf{h}_{p})/\tau)}
{\sum_{j=1}^{K}\exp(\text{sim}(\mathbf{h}_{anchor},\mathbf{h}_{j})/\tau)}.
\end{equation}
Using text $\mathbf{h}_{tl}$ as anchor, we define
\begin{align}
\mathcal{L}_{\text{text}} &= \mathcal{L}_{\text{InfoNCE}}(\mathbf{h}_{tl},\mathbf{h}_{tm}),\\
\mathcal{L}_{\text{vis}} &= \mathcal{L}_{\text{InfoNCE}}(\mathbf{h}_{tl},\mathbf{h}_{v}),\\
\mathcal{L}_{\text{aud}} &= \mathcal{L}_{\text{InfoNCE}}(\mathbf{h}_{tl},\mathbf{h}_{a}),\\
\mathcal{L}_{\text{fine}} &= \mathcal{L}_{\text{InfoNCE}}(\mathbf{h}_{tl},\mathbf{h}_{f}),
\end{align}
and sum them up:
\begin{equation}
\mathcal{L}_{\text{contrastive}}
=\mathcal{L}_{\text{text}}+\mathcal{L}_{\text{vis}}
+\mathcal{L}_{\text{aud}}+\mathcal{L}_{\text{fine}}.
\end{equation}

\subsection{Classification Objective}
\vspace{-0.5em}
The classifier is trained with cross-entropy over logits from $M_{cf}$:
\begin{equation}
\mathcal{L}_{\text{cls}}=\text{CrossEntropy}(WM_{cf}+b,y).
\end{equation}
The final loss combines all objectives:
\begin{equation}
\mathcal{L}=\mathcal{L}_{\text{cls}}+\mathcal{L}_{\text{proto}}
+\mathcal{L}_{\text{contrastive}}.
\end{equation}
\begin{table*}[t]
\centering
\setlength{\tabcolsep}{6pt}
\renewcommand{\arraystretch}{1.1}
\caption{Ablation Study over different loss functions on MIntRec\cite{b5} and MIntRec2.0\cite{b6}. The best performance for each metric is highlighted in bold.}
\label{tab:loss-ablation}
\resizebox{\linewidth}{!}{
\begin{tabular}{ccc|cccc|cccc}
\toprule
\multicolumn{3}{c|}{\textbf{Loss Functions}} &
\multicolumn{4}{c|}{\textbf{MIntRec\cite{b5}}} &
\multicolumn{4}{c}{\textbf{MIntRec2.0\cite{b6}}} \\
\cmidrule(lr){1-3}\cmidrule(lr){4-7}\cmidrule(lr){8-11}
Classifier & Contrastive & Prototype &
ACC (\%) & WF1 (\%) & WP (\%) & R (\%) &
ACC (\%) & WF1 (\%) & WP (\%) & R (\%) \\
\midrule
$\checkmark$ & $\times$ & $\times$ &
74.16 & 73.88 & 73.99 & 71.30 &
59.47 & 58.49 & 58.32 & 52.37 \\
$\checkmark$ & $\checkmark$ & $\times$ &
74.61 & 74.42 & 75.55 & 71.87 &
60.11 & 58.74 & 59.32 & 52.53 \\
$\checkmark$ & $\times$ & $\checkmark$ &
74.83 & 74.60 & 74.97 & 71.46 &
60.01 & 58.96 & 59.49 & 52.71 \\
$\checkmark$ & $\checkmark$ & $\checkmark$ &
\textbf{76.18} & \textbf{75.66} & \textbf{76.17} & \textbf{74.39} &
\textbf{60.40} & \textbf{59.23} & \textbf{60.51} & \textbf{53.96} \\
\bottomrule
\end{tabular}}
\end{table*}
% \vspace{-2.5em}
\section{Experiments}
\label{sec:experiments}
% \vspace{-0.5em}
\subsection{Datasets}
% \vspace{-0.5em}
We conduct experiments on two widely used benchmarks for multimodal intent recognition: \textbf{MIntRec 1.0}~\cite{b5} contains 2,224 high-quality samples with 20 intent classes, collected from multimodal human-machine interactions. Each instance is composed of a textual query, an aligned video segment, and its corresponding acoustic signal. \textbf{MIntRec2.0}~\cite{b6} extends MIntRec 1.0 to a more challenging setting with 30 intent classes, incorporating long-tailed distributions and open-intent utterances. This dataset encompasses 1,245 high-quality dialogues, aggregating a total of 15,040 samples that incorporate text, video, and audio modalities.
\vspace{-1em}
\subsection{Baselines and Experiment Setting}
\vspace{-0.5em}
We compare our proposed MVCL-DAF++ with the following competitive baselines: 
\textbf{MulT}~\cite{b7}, \textbf{MAG-BERT}~\cite{b19}, \textbf{TCL-MAP}~\cite{b11}, and \textbf{MVCL-DAF}~\cite{b18}. 
These baselines represent the state-of-the-art in multimodal intent recognition under both aligned and unaligned settings. 
All models are trained with AdamW (learning rate $2{\times}10^{-5}$, weight decay $0.2$, batch size $32$, and temperature $\tau{=}0.1$) for up to 100 epochs with early stopping of 10 epochs, using a single NVIDIA A100-40GB GPU. 
The reported results are averaged over 10 random seeds to ensure statistical robustness.

\subsection{Evaluation Metrics}
To comprehensively evaluate model performance, we adopt four widely used classification metrics: 
Accuracy (Acc), Recall, Weighted Precision, and Weighted F1-score. 

Accuracy measures the overall proportion of correctly predicted samples. 
Recall evaluates the model’s ability to correctly identify instances of each intent class, 
while Precision reflects the reliability of predicted labels. 
The Weighted Precision and Weighted F1-score are computed by weighting each class 
according to its sample frequency, thereby accounting for class imbalance 
and providing a more representative evaluation across all intent categories. 

Among these metrics, the Weighted F1-score serves as the primary evaluation criterion, 
as it balances both Precision and Recall while mitigating the influence of skewed class distributions.

% \vspace{-0.5em}
\section{Discussion}
\label{sec:discussion}
% \vspace{-0.5em}
\subsection{Performance Evaluation}
% \vspace{-0.5em}
OOur experimental results are summarized in Table~1, which shows that 
MVCL-DAF++ consistently outperforms all strong baselines across both 
the MIntRec and MIntRec2.0 datasets. 

On MIntRec, MVCL-DAF++ achieves an accuracy of 76.18\%, a weighted F1 score 
of 75.66\%, a weighted precision of 76.17\%, and a recall of 74.39\%, 
surpassing the previous best-performing method by clear margins on all 
evaluation metrics. These results indicate that the proposed prototype-aware 
contrastive alignment and coarse-to-fine fusion mechanism effectively enhance 
both classification reliability and cross-modal representation quality. 

On the more challenging MIntRec2.0 benchmark, which contains more ambiguous 
samples and stronger class imbalance, MVCL-DAF++ further demonstrates its 
robustness by achieving a weighted F1 score of 59.23\% and a recall of 53.96\%, 
substantially outperforming existing approaches. In particular, the performance 
gains in recall suggest that our method is better at identifying difficult or 
underrepresented intent categories, highlighting its improved generalization 
under realistic multimodal conditions. 

Overall, the consistent improvements across accuracy, precision, recall, and 
weighted F1 score on both datasets validate the effectiveness of our framework 
for robust multimodal intent recognition.

\begin{figure}[t]
    % \vspace{-2em}
    \centering
    \includegraphics[width=\textwidth, trim=80 180 175 110,clip,height=0.50625\linewidth, keepaspectratio]{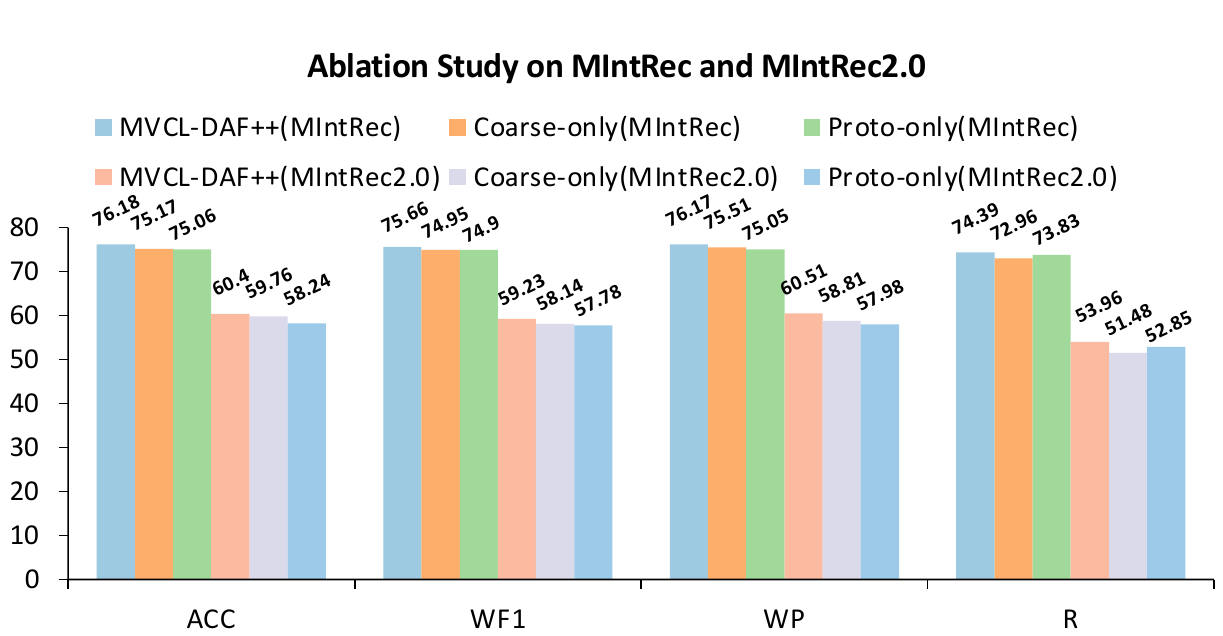}
    % \vspace{-1em}
    \caption{Ablation study on MIntRec\cite{b5} and MIntRec2.0\cite{b6}.
    }
    % \vspace{-1.5em}
    \label{fig:ablation}
\end{figure}
% \vspace{-1em}
\subsection{Ablation Study}
% \vspace{-0.5em}
To evaluate the effectiveness of each component and training objective in \textbf{MVCL-DAF++}, we conduct ablation studies on both MIntRec and MIntRec2.0. 
 We compare the full model with two variants: (1) \textit{Coarse-only}, which removes the prototype-aware contrastive alignment and keeps only the coarse-to-fine fusion; and (2) \textit{Proto-only}, which discards the coarse-to-fine fusion while retaining prototype alignment. As shown in Fig.~\ref{fig:ablation}, removing either module results in consistent performance drops across all metrics, confirming that both prototype-aware contrastive alignment and coarse-to-fine fusion are indispensable. For instance, on MIntRec, accuracy decreases from 76.18\% to 75.17\% and 75.06\% when removing the prototype or fusion, respectively. Similar trends are observed on the more challenging MIntRec2.0 benchmark, where the absence of either component reduces WF1 by average than 1.27 points.
We further examine the role of different loss functions (Table~\ref{tab:loss-ablation}). Using only the classification loss (Classifier) yields poor generalization, especially on MIntRec2.0 (59.47\% ACC). Adding contrastive loss (Contrastive) substantially improves robustness (60.11\% ACC), while prototype-aware contrastive alignment (Prototype) also provides notable gains. The best performance is achieved when combining all three losses, reaching 76.18\% ACC on MIntRec and 60.40\% ACC on MIntRec2.0. These results demonstrate that prototype-aware and contrastive objectives provide complementary benefits to classification, particularly under noisy and imbalanced conditions.
Overall, the ablation results highlight that both proposed modules and loss designs are essential for the robustness and generalizability of MVCL-DAF++.
% \vspace{-1em}
\begin{figure}[htp]
    \centering
    \includegraphics[width=\textwidth,trim=70 60 70 70,clip, height=0.50625\linewidth, keepaspectratio]{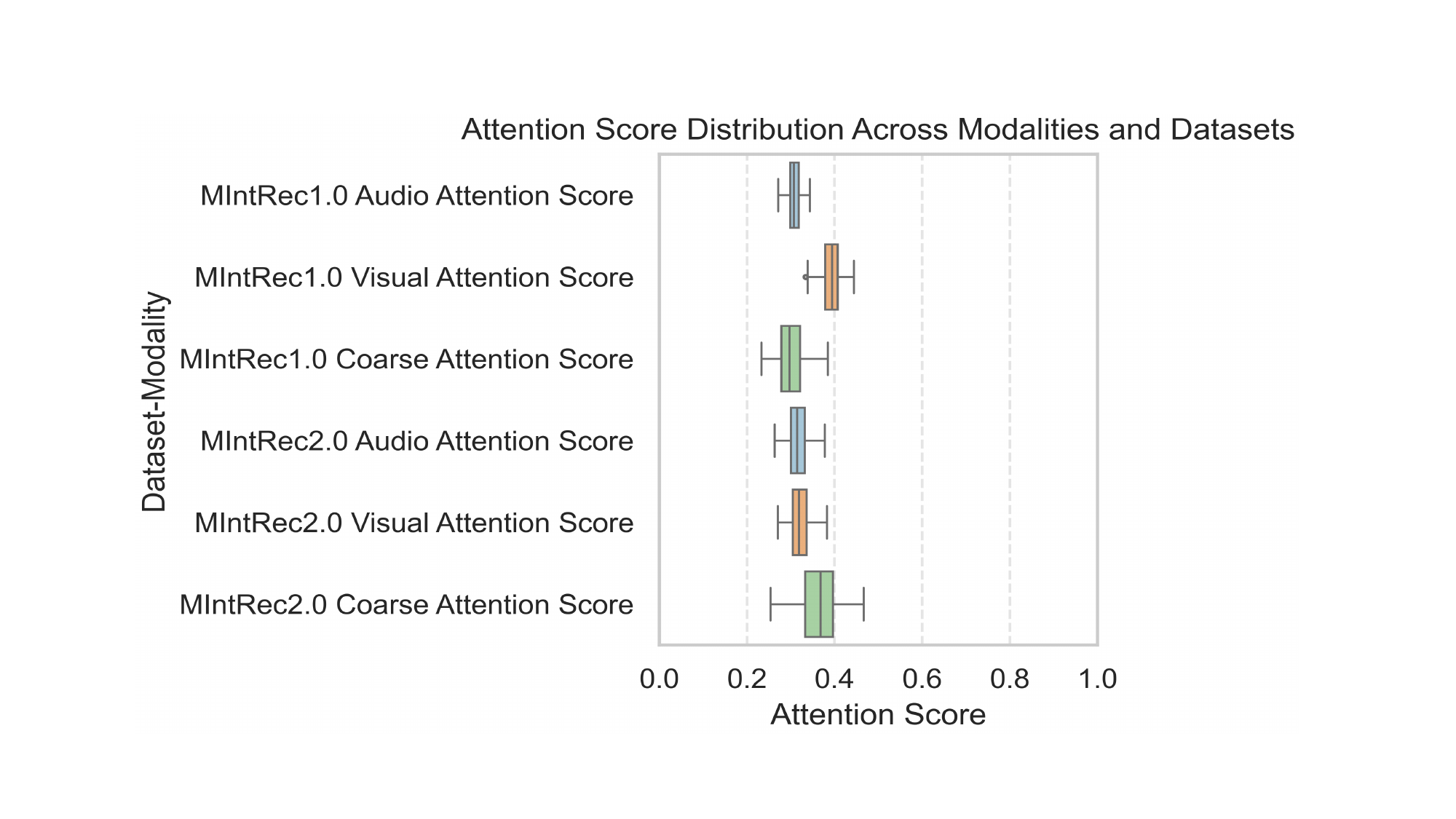}
    % \vspace{-2em}
    \caption{Distribution of attention scores across modalities and datasets.}
    \label{fig:coarse_attn_boxplot}
\end{figure}
% \vspace{-1.5em}
\begin{figure}[htp]
  \centering
  % \vspace{-1.5em}
  \includegraphics[width=\textwidth,trim=100 100 100 50,clip, height=0.50625\linewidth, keepaspectratio]{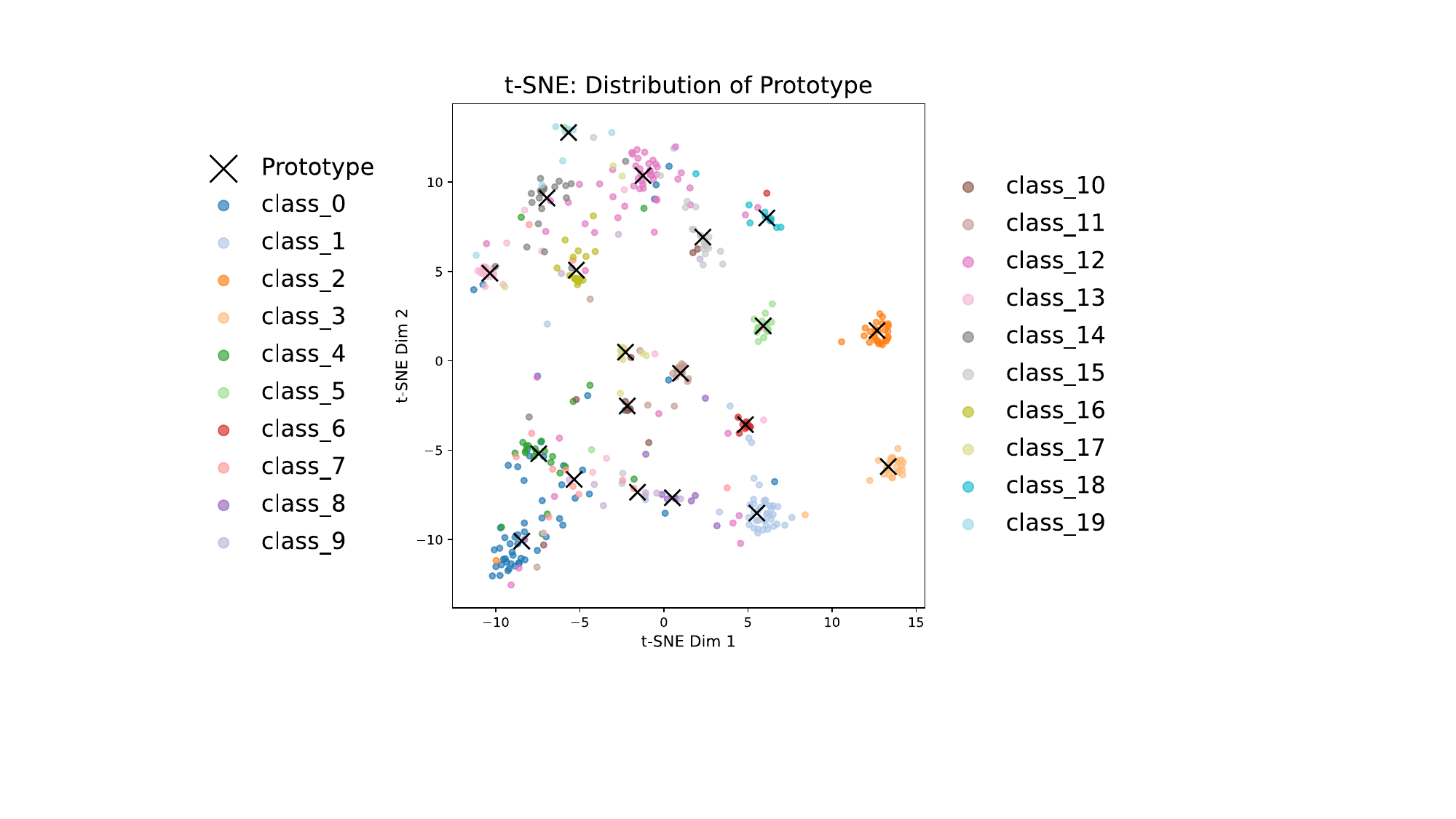}
  % \vspace{-1.5em}
  \caption{t-SNE visualization of learned embeddings (dots) and class prototypes (black crosses).}
  % \vspace{-1.5em}
  \label{fig:tsne_proto}
\end{figure}
% \vspace{-2em}
\subsection{Model Analysis}
% \vspace{-0.5em}
To further understand the inner mechanisms of our model, we analyze the modality attention and the prototype distributions. Fig.~\ref{fig:coarse_attn_boxplot} compares the attention scores assigned to the visual, acoustic, and coarse distributions within the coarse-to-fine fusion module. On the \textbf{MIntRec2.0} dataset, the model assigns relatively higher attention weights to coarse features, suggesting that in noisy and diverse environments, global cross-modal semantics are more reliable than token-level inputs. In contrast, for the cleaner \textbf{MIntRec} dataset, the attention weights are more evenly distributed between the modalities, indicating the sufficiency of fine-grained signals for intent inference.
To assess the effectiveness of prototype-aware contrastive learning, we visualize the learned embeddings and class prototypes using t-SNE in Fig.~\ref{fig:tsne_proto}. Each dot represents a sample, colored by class label, while black crosses indicate the corresponding prototypes. We observe that samples belonging to the same class are well clustered around their prototype anchors, confirming that the model has learned semantically consistent representations. Moreover, inter-class separability is enhanced, contributing to more discriminative alignment across modalities.
These analyses demonstrate that (1) the coarse-to-fine dynamic attention fusion mechanism enables context-aware fusion of modality cues, and (2) prototype-aware contrastive alignment promotes robust, label-consistent feature learning, both of which are crucial for generalization in real-world MMIR scenarios.

% \vspace{-2em}
\section{Conclusion}
\label{sec:conclusion}
% \vspace{-1em}
In this paper, we propose \textbf{MVCL-DAF++}, an enhanced MMIR framework that integrates prototype-aware contrastive alignment and coarse-to-fine dynamic attention fusion. By incorporating class-level semantic prototypes and leveraging global modality summaries alongside fine-grained features, our method achieves stronger semantic consistency and robustness. Extensive experiments on the MIntRec and MIntRec2.0 benchmarks demonstrate that MVCL-DAF++ significantly outperforms previous state-of-the-art models. Visualization and ablation studies further validate the effectiveness of our method. In future work, we plan to explore integrating this framework with large language model (LLM)-based architectures to enable more scalable and general multimodal intent recognition.

\section{Generative AI Use Disclosure}
Generative AI tools were used solely for language editing and manuscript polishing to improve clarity and readability. The technical content, model design, experimental setup, analysis, and conclusions were entirely developed by the authors. No generative AI system was involved in producing the scientific contributions of this work. All authors take full responsibility and accountability for the content of the paper and consent to its submission. No generative AI tool is listed as a co-author.
\section{Acknowledgements}
This work was supported by the Shenzhen Medical Research Fund 
(No.\ D2404001), and in part by the National Key Laboratory of the 
CAS on Medical Imaging Science and Technology System, the Key Research 
and Development Program of Guangdong Province (No.\ 2025B1111020001), 
the Shenzhen STIB programs (Nos.\ CJGJZD20230724093303007 and 
KJZD20240903101259001), and the Xisike Clinical Oncology Research 
Foundation (Y-2024AZ(NSCLC)MS-0156).

\bibliographystyle{IEEEtran}
\bibliography{ref}

@article{b1,
  author    = {Zhao, Y. and Wang, L. and Chen, H.},
  title     = {Deep learning approaches for multimodal intent recognition: A survey},
  journal   = {arXiv preprint arXiv:2507.22934},
  year      = {2025}
}

@inproceedings{b5,
  author    = {Zhang, H. and Xu, H. and Wang, X. and Zhou, Q. and Zhao, S. and Teng, J.},
  title     = {MIntRec: A new dataset for multimodal intent recognition},
  booktitle = {Proceedings of the ACM International Conference on Multimedia (MM)},
  year      = {2022},
  pages     = {1688--1697}
}

@inproceedings{b6,
  author    = {Zhang, H. and others},
  title     = {MIntRec2.0: A large-scale benchmark dataset for multimodal intent recognition and out-of-scope detection in conversations},
  booktitle = {Proceedings of the International Conference on Learning Representations (ICLR)},
  year      = {2024}
}

@inproceedings{b7,
  author    = {Tsai, Y.-H. H. and others},
  title     = {Multimodal Transformer for unaligned multimodal language sequences},
  booktitle = {Proceedings of the ACL},
  year      = {2019},
  pages     = {6558--6569}
}

@article{b8,
  author    = {Han, W. and Chen, H. and Poria, S.},
  title     = {Improving multimodal fusion with hierarchical mutual information maximization},
  journal   = {arXiv preprint arXiv:2109.00412},
  year      = {2021}
}

@inproceedings{b10,
  author    = {Xue, Z. and Marculescu, R.},
  title     = {Dynamic multimodal fusion},
  booktitle = {Proceedings of the CVPR Workshop on Multi-Modal Learning and Applications (MULA)},
  year      = {2023}
}

@inproceedings{b11,
  author    = {Zhou, Q. and Xu, H. and Li, H. and Zhang, H. and Zhang, X. and Wang, Y. and Gao, K.},
  title     = {Token-level contrastive learning with modality-aware prompting for multimodal intent recognition},
  booktitle = {Proceedings of the AAAI Conference on Artificial Intelligence},
  volume    = {38},
  year      = {2024}
}

@inproceedings{b12,
  author    = {Snell, J. and Swersky, K. and Zemel, R.},
  title     = {Prototypical networks for few-shot learning},
  booktitle = {Advances in Neural Information Processing Systems (NeurIPS)},
  year      = {2017}
}

@article{b13,
  author    = {Liang, P. and others},
  title     = {Factorized contrastive learning: Going beyond multi-view redundancy},
  journal   = {arXiv preprint arXiv:2306.05268},
  year      = {2023}
}

@article{b14,
  author    = {van den Oord, A. and Li, Y. and Vinyals, O.},
  title     = {Representation learning with contrastive predictive coding},
  journal   = {arXiv preprint arXiv:1807.03748},
  year      = {2019}
}

@inproceedings{b17,
  author    = {Sun, K. and Xie, Z. and Ye, M. and Zhang, H.},
  title     = {Contextual augmented global contrast for multimodal intent recognition},
  booktitle = {Proceedings of the IEEE/CVF Conference on Computer Vision and Pattern Recognition (CVPR)},
  year      = {2024},
  pages     = {26963--26973}
}

@inproceedings{b18,
  author    = {Hu, B. and Zhang, K. and Zhang, Y. and Ye, Y.},
  title     = {Adaptive multimodal fusion: Dynamic attention allocation for intent recognition},
  booktitle = {Proceedings of the AAAI Conference on Artificial Intelligence (AAAI)},
  year      = {2025}
}

@inproceedings{b19,
  author    = {Hasan, M. K. and Rahman, M. T. and Akhtar, M. S. and Ekbal, A. and Bhattacharyya, P.},
  title     = {MAG-BERT: Multimodal adaptation gate BERT for multimodal sentiment analysis},
  booktitle = {Proceedings of the International Conference on Computational Linguistics (COLING)},
  year      = {2020},
  pages     = {3615--3626}
}

@inproceedings{b20,
  author    = {Khosla, P. and Teterwak, P. and Wang, C. and Sarna, A. and Tian, Y. and Isola, P. and Maschinot, A. and Liu, C. and Krishnan, D.},
  title     = {Supervised contrastive learning},
  booktitle = {Advances in Neural Information Processing Systems (NeurIPS)},
  volume    = {33},
  year      = {2020},
  pages     = {18661--18673}
}

@inproceedings{b21,
  author    = {Yu, S. and Wang, Y. and Lin, Z. and Morency, L.-P.},
  title     = {Multimodal transformer with multi-scale alignment for multimodal sentiment analysis},
  booktitle = {Proceedings of ACL},
  year      = {2021}
}

@article{b22,
  title={HiCLIP: Contrastive Language-Image Pretraining with Hierarchy-aware Attention},
  author={Geng, Shijie and Yuan, Jianbo and Tian, Yu and Chen, Yuxiao and Zhang, Yongfeng},
  journal={arXiv preprint arXiv:2303.02995},
  year={2023}
}

@inproceedings{b23,
  title={Attention Bottlenecks for Multimodal Fusion},
  author={Nagrani, Arsha and Yang, Shan and Arnab, Anurag and Jansen, Aren and Schmid, Cordelia and Sun, Chen},
  booktitle={ICLR},
  year={2021}
}

@inproceedings{b24,
  author={Wang, Tongzhou and Isola, Phillip},
  title={Understanding Contrastive Representation Learning through Alignment and Uniformity on the Hypersphere},
  booktitle={Proceedings of the 37th International Conference on Machine Learning (ICML)},
  year={2020},
  pages={9929--9939}
}

@inproceedings{b25,
  author={Robinson, Joshua and Chuang, Ching-Yao and Sra, Suvrit and Jegelka, Stefanie},
  title={Can Contrastive Learning Avoid Shortcut Solutions?},
  booktitle={Advances in Neural Information Processing Systems (NeurIPS)},
  year={2021},
  pages={4974--4986}
}

@article{li2024towards,
  title={Towards visual-prompt temporal answer grounding in instructional video},
  author={Li, Shutao and Li, Bin and Sun, Bin and Weng, Yixuan},
  journal={IEEE transactions on pattern analysis and machine intelligence},
  volume={46},
  number={12},
  pages={8836--8853},
  year={2024},
  publisher={IEEE}
}

@inproceedings{diao2025temporal,
  title={Temporal Working Memory: Query-Guided Segment Refinement for Enhanced Multimodal Understanding},
  author={Diao, Xingjian and Zhang, Chunhui and Wu, Weiyi and Ouyang, Zhongyu and Qing, Peijun and Cheng, Ming and Vosoughi, Soroush and Gui, Jiang},
  booktitle={Findings of the Association for Computational Linguistics: NAACL 2025},
  pages={3393--3409},
  year={2025}
}

@article{yu2025visual,
  title={Visual Document Understanding and Question Answering: A Multi-Agent Collaboration Framework with Test-Time Scaling},
  author={Yu, Xinlei and Chen, Zhangquan and Zhang, Yudong and Lu, Shilin and Shen, Ruolin and Zhang, Jiangning and Hu, Xiaobin and Fu, Yanwei and Yan, Shuicheng},
  journal={arXiv preprint arXiv:2508.03404},
  year={2025}
}

@article{zhu2025survey,
  title={A Survey on Multi-modal Intent Recognition: Recent Advances and New Frontiers},
  author={Zhu, Zhihong and Zhang, Fan and Zhang, Yunyan and Sun, Jinghan and Huang, Zhiqi and Long, Qingqing and Xing, Bowen and Wu, Xian},
  journal={Findings of the Association for Computational Linguistics: EMNLP 2025},
  pages={15223--15236},
  year={2025}
}

\end{document}